\documentclass[runningheads]{llncs}
\usepackage{graphicx}
\usepackage{amsmath,amssymb} \usepackage{color}

\usepackage[numbers,sort&compress]{natbib}

\usepackage{wrapfig}
\usepackage{array}
\usepackage{multirow}

\usepackage[usenames,dvipsnames]{xcolor}
\usepackage{paralist}
\usepackage{xspace}
\usepackage{comment}

\makeatletter
\let\MYcaption\@makecaption
\makeatother

\usepackage[font=footnotesize]{subcaption}
\makeatletter
\let\@makecaption\MYcaption
\makeatother

\graphicspath{{./figure/}}

\newcommand{\etal}[0]{\textit{et~al.}\xspace}
\newcommand{\figref}[1]{Figure~\ref{#1}}

\newcommand{\vvec}{\ensuremath\vec{v}}
\newcommand{\gvec}{\ensuremath\vec{g}}
\newcommand{\gmap}{\ensuremath\vec{m}}
\newcommand{\x}{\ensuremath\vec{x}}
\newcommand{\abs}[1]{\ensuremath\left|#1\right|}

\usepackage{hyperref}

\begin{document}
\title{Deep Pictorial Gaze Estimation}
\titlerunning{Deep Pictorial Gaze Estimation}

\author{  Seonwook Park \and  Adrian Spurr \and  Otmar Hilliges }
\institute{AIT Lab, Department of Computer Science, ETH Zurich\\
\texttt{\{firstname.lastname\}@inf.ethz.ch}
}

\authorrunning{S. Park et al.}

\institute{AIT Lab, Department of Computer Science, ETH Zurich
\email{\{firstname.lastname\}@inf.ethz.ch}
}
\maketitle

\begin{abstract}
Estimating human gaze from natural eye images only is a challenging task. Gaze direction can be defined by the pupil- and the eyeball center where the latter is unobservable in 2D images. Hence, achieving highly accurate gaze estimates is an ill-posed problem.
In this paper, we introduce a novel deep neural network architecture specifically designed for the task of gaze estimation from single eye input.
Instead of directly regressing two angles for the pitch and yaw of the eyeball, we regress to an intermediate pictorial representation which in turn simplifies the task of 3D gaze direction estimation.
Our quantitative and qualitative results show that our approach achieves higher accuracies than the state-of-the-art and is robust to variation in gaze, head pose and image quality.

\end{abstract}

\begin{figure}[t]
	\includegraphics[width=\columnwidth]{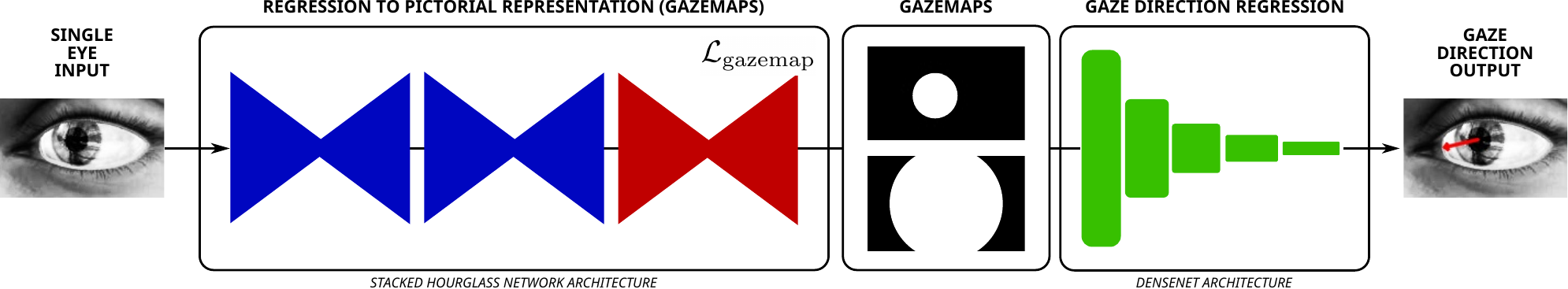}
	\caption{Our sequential neural network architecture first estimates a novel pictorial representation of 3D gaze direction, then performs gaze estimation from the minimal image representation to yield improved performance on MPIIGaze, Columbia and EYEDIAP.}
	\label{fig:architecture}
\end{figure}

\section{Introduction}
Accurately estimating human gaze direction has many applications in assistive technologies for users with motor disabilities \cite{Chin2008JRRD}, gaze-based human-computer interaction \cite{majaranta14_apc}, visual attention analysis \cite{Liu2011}, consumer behavior research \cite{Wedel2008}, AR, VR and more.
Traditionally this has been done via specialized hardware, shining infrared illumination into the user's eyes and via specialized cameras, sometimes requiring use of a headrest.
Recently deep learning based approaches have made first steps towards fully unconstrained gaze estimation under free head motion, in environments with uncontrolled illumination conditions, and using only a single commodity (and potentially low quality) camera. However, this remains a challenging task due to inter-subject variance in eye appearance, self-occlusions, and head pose and rotation variations. In consequence, current approaches attain accuracies in the order of $6^\circ$ only and are still far from the requirements of many application scenarios.
While demonstrating the feasibility of purely image based gaze estimation and introducing large datasets, these learning-based approaches \cite{Zhang2015CVPR,Krafka2016CVPR,Zhang2017CVPRW} have leveraged convolutional neural network (CNN) architectures, originally designed for the task of image classification, with minor modifications.
For example, \cite{Zhang2015CVPR,Zhang2017PAMI} simply append head pose orientation to the first fully connected layer of either LeNet-5 or VGG-16, while \cite{Krafka2016CVPR} proposes to merge multiple input modalities by replicating convolutional layers from AlexNet.
In \cite{Zhang2017CVPRW} the AlexNet architecture is modified to learn so-called spatial-weights to emphasize important activations by region when full face images are provided as input.
Typically, the proposed architectures are only supervised via a mean-squared error loss on the gaze direction output, represented as either a 3-dimensional unit vector or pitch and yaw angles in radians.

In this work we propose a network architecture that has been specifically designed with the task of gaze estimation in mind.
An important insight is that regressing first to an abstract but gaze specific representation helps the network to more accurately predict the final output of 3D gaze direction.
Furthermore, introducing this gaze representation also allows for intermediate supervision which we experimentally show to further improve accuracy.
Our work is loosely inspired by recent progress in the field of human pose estimation. Here, earlier work directly regressed joint coordinates \cite{Toshev2014CVPR}. More recently the need for a more task specific form of supervision has led to the use of confidence maps or heatmaps, where the position of a joint is depicted as a 2-dimensional Gaussian \cite{Tompson2014NIPS,Newell2016ECCV,Wei2016CVPR}.
This representation allows for a simpler mapping between input image and joint position, allows for intermediate supervision, and hence for deeper networks.
However, applying this concept of heatmaps to regularize training is not directly applicable to the case of gaze estimation since the crucial eyeball center is not observable in 2D image data.
We propose a conceptually similar representation for gaze estimation, called \emph{gazemaps}. Such a gazemap is an abstract, pictorial representation of the eyeball, the iris and the pupil at it's center (see \figref{fig:architecture}).

The simplest depiction of an eyeball's rotation can be made via a circle and an ellipse, the former representing the eyeball, and the latter the iris. The gaze direction is then defined by the vector connecting the larger circle's center and the ellipse.
Thus 3D gaze direction can be (pictorially) represented in the form of an image, where a spherical eyeball and circular iris are projected onto the image plane, resulting in a circle and ellipse. Hence, changes in gaze direction result in changes in ellipse positioning (cf. \figref{fig:gmap_diagram}).
This pictorial representation can be easily generated from existing training data, given known gaze direction annotations.
At inference time recovering gaze direction from such a pictorial representation is a much simpler task than regressing directly from raw pixel values.
However, adapting the input image to fit our pictorial representation is non-trivial.
For a given eye image, a circular eyeball and an ellipse must be fitted, then centered and rescaled to be in the expected shape.
We experimentally observed that this task can be performed well using a fully convolutional architecture. Furthermore, we show that our approach outperforms prior work on the final task of gaze estimation significantly.

Our main contribution consists of a novel architecture for appearance-based gaze estimation. At the core of the proposed architecture lies the pictorial representation of 3D gaze direction to which the network fits the raw input images and from which additional convolutional layers estimate the final gaze direction.
In addition, we perform:
\begin{inparaenum}[(a)]
  \item an in-depth analysis of the effect of intermediate supervision using our pictorial representation,
  \item quantitative evaluation and comparison against state-of-the-art gaze estimation methods on three challenging datasets (MPIIGaze, EYEDIAP, Columbia) in the person independent setting, and a
  \item detailed evaluation of the robustness of a model trained using our architecture in terms of gaze direction and head pose as well as image quality.
\end{inparaenum}
Finally, we show that our method reduces gaze error by $18\%$ compared to the state-of-the-art \cite{Zhang2017PAMI} on MPIIGaze.

\section{Related Work}
Here we briefly review the most important work in eye gaze estimation and review work touching on relevant aspects in terms of network architecture from adjacent areas such as image classification and human pose estimation.

\subsection{Appearance-based Gaze Estimation with CNNs}
Traditional approaches to image-based gaze estimation are typically categorized as \emph{feature-based} or \emph{model-based}. Feature-based approaches reduce an eye image down to a set of features based on hand-crafted rules \cite{Sesma2012ETRA,Huang2014MM,Wood2014ETRAEA,Huang2017MVA} and then feed these features into simple, often linear machine learning models to regress the final gaze estimate. Model-based methods instead attempt to fit a known 3D model to the eye image \cite{Xiong2014Ubicomp,Sun2015Elsevier,Wood2016ECCV,Wang2017ICCV} by minimizing a suitable energy.

Appearance-based methods learn a direct mapping from raw eye images to gaze direction.
Learning this direct mapping can be very challenging due to changes in illumination, (partial) occlusions, head motion and eye decorations. Due to these challenges, appearance-based gaze estimation methods required the introduction of large, diverse training datasets and typically leverage some form of convolutional neural network architecture.

Early works in appearance-based methods were restricted to laboratory settings with fixed head pose \cite{Baluja1994,Tan2002WACV}.
These initial constraints have become progressively relaxed, notably by the introduction of new datasets collected in everyday settings \cite{Zhang2015CVPR,Krafka2016CVPR} or in simulated environments \cite{Sugano2014CVPR,Wood2015ICCV,Wood2016ETRA}.
The increasing scale and complexity of training data has given rise to a wide variety of learning-based methods including variations of linear regression \cite{Lu2011BMVC,Lu2011ICCV,FunesMora2014ETRA}, random forests \cite{Sugano2014CVPR}, $k$-nearest neighbours \cite{Sugano2014CVPR,Wood2016ETRA}, and CNNs \cite{Zhang2015CVPR,Krafka2016CVPR,Zhang2017PAMI,Zhang2017CVPRW,Wood2015ICCV,Shrivastava2017CVPR}.
CNNs have proven to be more robust to visual appearance variations, and are capable of person-independent gaze estimation when provided with sufficient scale and diversity of training data.
Person-independent gaze estimation can be performed without a user calibration step, and can directly be applied to areas such as visual attention analysis on unmodified devices \cite{Papoutsaki16IJCAI}, interaction on public displays \cite{Zhang13CHI}, and identification of gaze targets \cite{Zhang17UIST}, albeit at the cost of increased need for training data and computational cost.

Several CNN architectures have been proposed for person-independent gaze estimation in unconstrained settings, mostly differing in terms of possible input data modalities.
Zhang \etal \cite{Zhang2015CVPR,Zhang2017CVPRW} adapt the LeNet-5 and VGG-16 architectures such that head pose angles (pitch and yaw) are concatenated to the first fully-connected layers.
Despite its simplicity this approach yields the current best gaze estimation error of $5.5^\circ$ when evaluating for the within-dataset cross-person case on MPIIGaze with single eye image and head pose input.
In \cite{Krafka2016CVPR} separate convolutional streams are used for left/right eye images, a face image, and a $25\times 25$ grid indicating the location and scale of the detected face in the image frame.
Their experiments demonstrate that this approach yields improvements compared to \cite{Zhang2015CVPR}.
In \cite{Zhang2017CVPRW} a single face image is used as input and so-called spatial-weights are learned. These emphasize important features based on the input image, yielding considerable improvements in gaze estimation accuracy.

We introduce a novel pictorial representation of eye gaze and incorporate this into a deep neural network architecture via intermediate supervision. To the best of our knowledge we are the first to apply fully convolutional architecture to the task of appearance-based gaze estimation. We show that together these contribution lead to a significant performance improvement of $18\%$ even when using a single eye image as sole input.
\subsection{Deep Learning with Auxiliary Supervision}
It has been shown \cite{Lee15AIStats,Szegedy15CVPR} that by applying a loss function on intermediate outputs of a network, better performance can be yielded in different tasks.
This technique was introduced to address the vanishing gradients problem during the training of deeper networks.
In addition, such intermediate supervision allows for the network to quickly learn an estimate for the final output then learn to refine the predicted features - simplifying the mappings which need to be learned at every layer.
Subsequent works have adopted intermediate supervision \cite{Wei2016CVPR,Newell2016ECCV} to good effect for human pose estimation, by replicating the final output loss.

Another technique for improving neural network performance is the use of auxiliary data through multi-task learning.
In \cite{Zhang14ECCV,Ranjan2016arXiv}, the architectures are formed of a single shared convolutional stream which is split into separate fully-connected layers or regression functions for the auxiliary tasks of gender classification, face visibility, and head pose.
Both works show marked improvements to state-of-the-art results in facial landmarks localization.
In these approaches through the introduction of multiple learning objectives, an implicit prior is forced upon the network to learn a representation that is informative to both tasks.
On the contrary, we \emph{explicitly} introduce a gaze-specific prior into the network architecture via gazemaps.

Most similar to our contribution is the work in \cite{Honari2018CVPR} where facial landmark localization performance is improved by applying an auxiliary emotion classification loss.
A key aspect to note is that their network is sequential, that is, the emotion recognition network takes only facial landmarks as input.
The detected facial landmarks thus act as a manually defined representation for emotion classification, and creates a bottleneck in the full data flow.
It is shown experimentally that applying such an auxiliary loss (for a different task) yields improvement over state-of-the-art results on the AFLW dataset.
In our work, we learn to regress an intermediate and minimal representation for gaze direction, forming a bottleneck before the main task of regressing two angle values.
Thus, an important distinction to \cite{Honari2018CVPR} is that while we employ an auxiliary loss term, it directly contributes to the task of gaze direction estimation.
Furthermore, the auxiliary loss is applied as an intermediate task.
We detail this further in Sec.~\ref{sec:representation}.

Recent work in multi-person human pose estimation \cite{Cao17CVPR} learns to estimate joint location heatmaps alongside so-called ``part affinity fields''.
When combined, the two outputs then enable the detection of multiple peoples' joints with reduced ambiguity in terms of which person a joint belongs to.
In addition, at the end of every image scale, the architecture concatenates feature maps from each separate stream such that information can flow between the ``part confidence'' and ``part affinity'' maps.
Thus, they operate on the image representation space, taking advantage of the strengths of convolutional neural networks.
Our work is similar in spirit in that it introduces a novel image-based representation.

\section{Method}
A key contribution of our work is a pictorial representation of 3D gaze direction - which we call \emph{gazemaps}.
This representation is formed of two boolean maps, which can be regressed by a fully convolutional neural network.
In this section, we describe our representation (Sec.~\ref{sec:representation}) then explain how we constructed our architecture to use the representation as reference for intermediate supervision during training of the network (Sec.~\ref{sec:architecture}).

\begin{figure}
\begin{minipage}{0.50\columnwidth}
	\centering
	\includegraphics[width=\columnwidth]{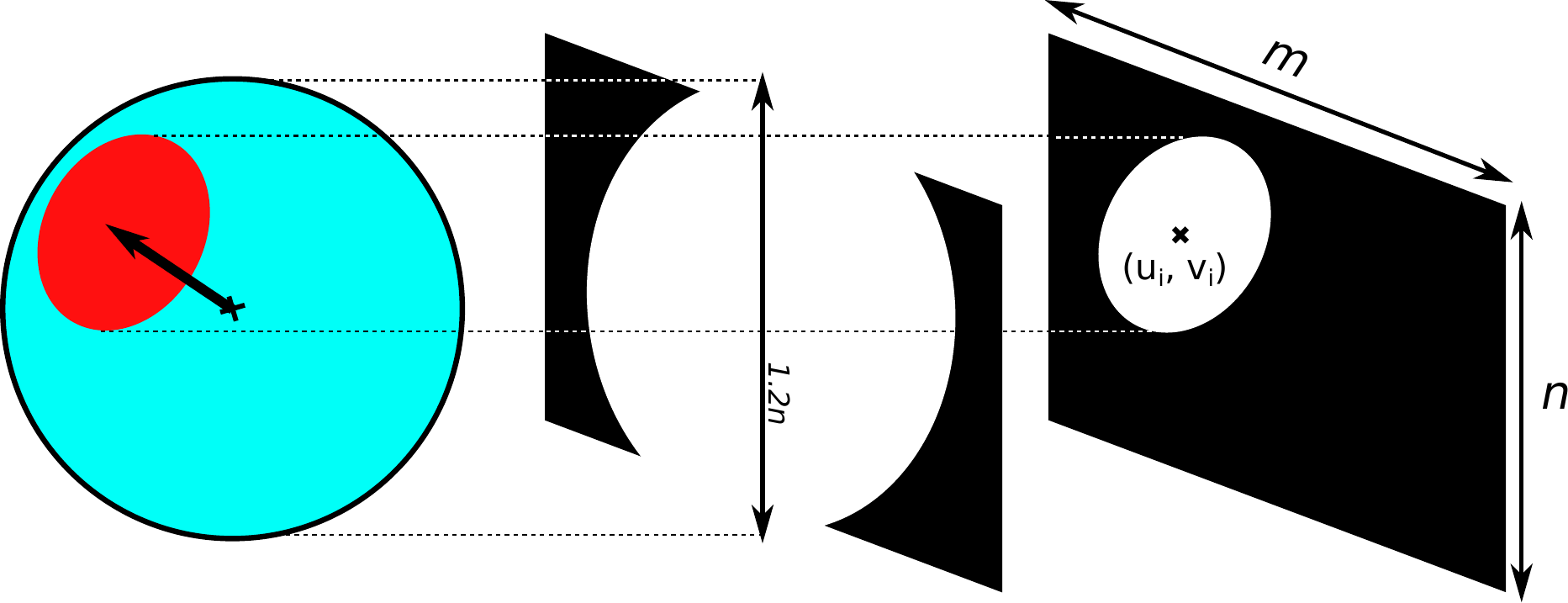}
	\subcaption{}
	\label{fig:gmap_diagram}
\end{minipage}
\hfill
\begin{minipage}{0.46\columnwidth}
	\centering
		\includegraphics[width=\columnwidth]{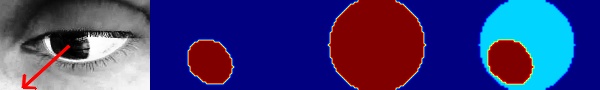}\\[0.2mm]
	\includegraphics[width=\columnwidth]{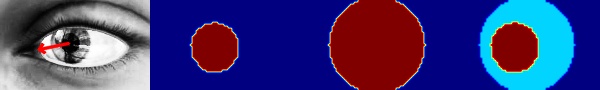}\\[0.2mm]
	\includegraphics[width=\columnwidth]{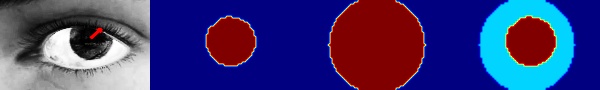}
	\subcaption{Example gazemaps from UnityEyes}
	\label{fig:gmap_samples}
\end{minipage}
\caption{Our pictorial representation of 3D gaze direction, essentially a projection of simple eyeball and iris models onto binary maps (a). Example-pairs are shown in (b) with (left-to-right) input image, iris map, eyeball map, and a superimposed visualization.}
\end{figure}

\subsection{Pictorial Representation of 3D Gaze}
\label{sec:representation}

In the task of appearance-based gaze estimation, an input eye image is processed to yield gaze direction in 3D.
This direction is often represented as a $3$-element unit vector $\vvec$ \cite{FunesMora2016IJCV,Zhang2017CVPRW,Shrivastava2017CVPR}, or as two angles representing eyeball pitch and yaw $\gvec = \left(\theta,\,\phi\right)$ \cite{Sugano2014CVPR,Wood2015ICCV,Zhang2015CVPR,Zhang2017PAMI}.
In this section, we propose an alternative to previous direct mappings to $\vvec$ or $\gvec$.

If we state the input eye images as $\vec{x}$ and regard regressing the values $\vec{g}$, a conventional gaze estimation model estimates $f: \x\rightarrow\gvec$.
The mapping $f$ can be complex, as reflected by the improvement in accuracies that have been attained by simple adoption of newer CNN architectures ranging from LeNet-5 \cite{Zhang2015CVPR,Shrivastava2017CVPR}, AlexNet \cite{Krafka2016CVPR,Zhang2017CVPRW}, to VGG-16 \cite{Zhang2017PAMI}, the current state-of-the-art CNN architecture for appearance-based gaze estimation.
We hypothesize that it is possible to learn an intermediate image representation of the eye, $\gmap$.
That is, we define our model as $\gvec = k \circ j(x)$ where $j: \x\rightarrow\gmap$ and $k:\gmap\rightarrow\gvec$.
It is conceivable that the complexity of learning $j$ and $k$ should be significantly lower than directly learning $f$, allowing for neural network architectures with significantly lower model complexity to be applied to the same task of gaze estimation with higher or equivalent performance.

Thus, we propose to estimate so-called \emph{gazemap}s ($\gmap$) and from that the 3D gaze direction ($\gvec$).
We reformulate the task of gaze estimation into two concrete tasks:
\begin{inparaenum}[(a)]
	\item reduction of input image to minimal normalized form (gazemaps),
	and
	\item gaze estimation from gazemaps.
\end{inparaenum}

The gazemaps for a given input eye image should be visually similar to the input yet distill only the necessary information for gaze estimation to ensure that the mapping $k: \gmap\rightarrow\gvec$ is simple.
To do this, we consider that an average human eyeball has a diameter of $\approx24mm$ \cite{Bekerman2014} while an average human iris has a diameter of $\approx12mm$ \cite{Forrester2015}. We then assume a simple model of the human eyeball and iris, where the eyeball is a perfect sphere, and the iris is a perfect circle.
For an output image dimension of $m\times n$, we assume the projected eyeball diameter $2r = 1.2n$ and calculate the iris centre coordinates $\left(u_i,\,v_i\right)$ to be:
\begin{align}
	u_i &= \frac{m}{2} - r^\prime \sin\phi \cos\theta\\
	v_i &= \frac{n}{2} - r^\prime \sin\theta
\end{align}
where $r^\prime = r \cos\left(\sin^{-1} \frac{1}{2}\right)$,
and gaze direction $\gvec=\left(\theta,\phi\right)$.
The iris is drawn as an ellipse with major-axis diameter of $r$ and minor-axis diameter of $r\abs{\cos\theta\cos\phi}$.
Examples of our gazemaps are shown in Fig.~\ref{fig:gmap_samples} where two separate boolean maps are produced for one gaze direction $\gvec$.

Learning how to predict gazemaps only from a single eye image is not a trivial task.
Not only do extraneous factors such as image artifacts and partial occlusion need to be accounted for, a simplified eyeball must be fit to the given image based on iris and eyelid appearance. The detected regions must then be scaled and centered to produce the gazemaps. Thus the mapping $j: \x\rightarrow\gmap$ requires a more complex neural network architecture than the mapping $k: \gmap\rightarrow\gvec$.

\subsection{Neural Network Architecture}
\label{sec:architecture}
Our neural network consists of two parts:
\begin{inparaenum}[(a)]
	\item regression from eye image to gazemap,
	and
	\item regression from gazemap to gaze direction $\gvec$.
\end{inparaenum}
While any CNN architecture can be implemented for (b), regressing (a) requires a fully convolutional architecture such as those used in human pose estimation.
We adapt the stacked hourglass architecture from Newell~\etal~\cite{Newell2016ECCV} for this task.
The hourglass architecture has been proven to be effective in tasks such as human pose estimation and facial landmarks detection \cite{Zafeiriou2017CVPRW} where complex spatial relations need to be modeled at various scales to estimate the location of occluded joints or key points.
The architecture performs repeated multi-scale refinement of feature maps, from which desired output confidence maps can be extracted via $1\times 1$ convolution layers.
We exploit this fact to have our network predict gazemaps instead of classical confidence or heatmaps for joint positions.
In Sec.~\ref{sec:evaluations}, we demonstrate that this works well in practice.

In our gazemap-regression network, we use $3$ hourglass modules with intermediate supervision applied on the gazemap outputs of the last module only.
The minimized intermediate loss is:
\begin{equation}
	\mathcal{L}_\mathrm{gazemap} = 	-\alpha \sum_{p\in\mathcal{P}} 		\gmap(p) \log \hat{\gmap}(p)  ,
        \label{eq:gazemap_loss}
\end{equation}
where
we calculate a cross-entropy between predicted $\hat{\gmap}$ and ground-truth gazemap $\gmap$ for pixels $p$ in set of all pixels $\mathcal{P}$.
In our evaluations, we set the weight coefficient $\alpha$ to $10^{-5}$.

For the regression to $\gvec$, we select DenseNet which has recently been shown to perform well on image classification tasks \cite{Huang2017CVPR} while using fewer parameters compared to previous architectures such as ResNet \cite{He2016CVPR}.
The loss term for gaze direction regression (per input) is:
\begin{equation}
	\mathcal{L}_\mathrm{gaze} = \abs{\abs{\gvec - \hat{\gvec}}}^2_2  ,
        \label{eq:gaze_loss}
\end{equation}
where $\tilde\gvec$ is the gaze direction predicted by our neural network.

\section{Implementation}
In this section, we describe the fully convolutional (Hourglass) and regressive (DenseNet) parts of our architecture in more detail.

\subsection{Hourglass Network}

In our implementation of the Stacked Hourglass Network \cite{Newell2016ECCV}, we provide images of size $150\times90$ as input, and refine $64$ feature maps of size $75\times45$ throughout the network.
The half-scale feature maps are produced by an initial convolutional layer with filter size $7$ and stride $2$ as done in the original paper \cite{Newell2016ECCV}.
This is followed by batch normalization, ReLU activation, and two residual modules before being passed as input to the first hourglass module.

There exist $3$ hourglass modules in our architecture, as visualized in Figure~\ref{fig:architecture}.
In human pose estimation, the commonly used outputs are 2-dimensional confidence maps, which are pixel-aligned to the input image.
Our task differs, and thus we do not apply intermediate supervision to the output of every hourglass module.
This is to allow for the input image to be processed at multiple scales over many layers, with the necessary features becoming aligned to the final output gazemap representation.
Instead, we apply $1\times1$ convolutions to the output of the last hourglass module, and apply the gazemap loss term (Eq.~\ref{eq:gazemap_loss}).

\begin{figure}
	\centering
	\includegraphics[width=0.8\columnwidth]{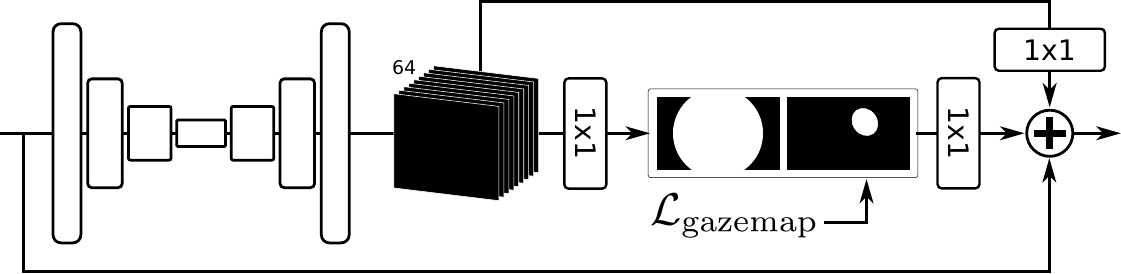}
	\caption{
		Intermediate supervision is applied to the output of an hourglass module by performing $1\times1$ convolutions.
		The intermediate gazemaps and feature maps from the previous hourglass module are then concatenated back into the network to be passed onto the next hourglass module as is done in the original Hourglass paper \cite{Newell2016ECCV}.}
\end{figure}
\vspace{-10pt}

\subsection{DenseNet}
\label{sec:densenet}
As described in Section~\ref{sec:representation}, our pictorial representation allows for a simpler function to be learnt for the actual task of gaze estimation.
To demonstrate this, we employ a very lightweight DenseNet architecture \cite{Huang2017CVPR}.
Our gaze regression network consists of $5$ dense blocks ($5$ layers per block) with a growth-rate of $8$, bottleneck layers, and a compression factor of $0.5$.
This results in just $62$ feature maps at the end of the DenseNet, and subsequently $62$ features through global average pooling.
Finally, a single linear layer maps these features to $\gvec$.
The resulting network is light-weight and consists of just $66$k trainable parameters.

\subsection{Training Details}
We train our neural network with a batch size of $32$, learning rate of $0.0002$ and $L_2$ weights regularization coefficient of $10^{-4}$.
The optimization method used is Adam \cite{Kingma2014arXiv}.
Training occurs for $20$ epochs on a desktop PC with an Intel Core i7 CPU and Nvidia Titan Xp GPU, taking just over $2$ hours for one fold (out of $15$) of a leave-one-person-out evaluation on the MPIIGaze dataset.

During training, slight data augmentation is applied in terms of image translation and scaling, and learning rate is multiplied by $0.1$ after every $5$k gradient update steps, to address over-fitting and to stabilize the final error.

\begin{figure}[t]
	\centering
	\begin{minipage}{\columnwidth}
		\includegraphics[width=0.083\columnwidth]{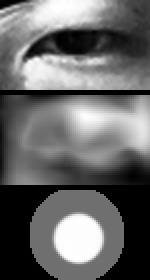} \hfill
		\includegraphics[width=0.083\columnwidth]{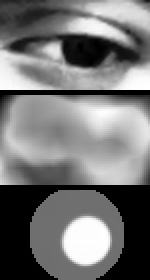} \hfill
		\includegraphics[width=0.083\columnwidth]{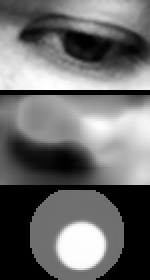} \hfill
		\includegraphics[width=0.083\columnwidth]{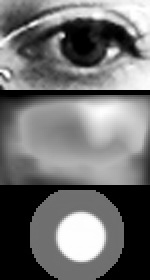} \hfill
		\includegraphics[width=0.083\columnwidth]{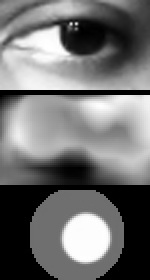} \hfill
		\includegraphics[width=0.083\columnwidth]{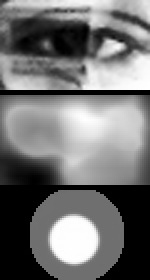} \hfill
		\includegraphics[width=0.083\columnwidth]{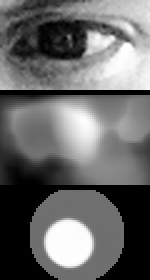} \hfill
		\includegraphics[width=0.083\columnwidth]{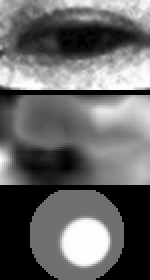} \hfill
		\includegraphics[width=0.083\columnwidth]{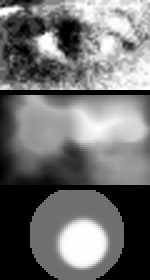} \hfill
		\includegraphics[width=0.083\columnwidth]{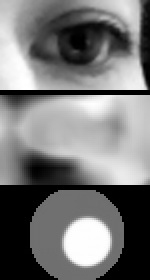} \hfill
		\includegraphics[width=0.083\columnwidth]{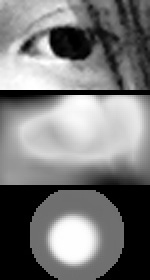} \hfill
				\subcaption{\scriptsize Intermediate representations of training samples without (middle) and with (bottom) intermediate supervision}\label{fig:gazemap_comp}
	\end{minipage}
	\vskip 1mm
	\begin{minipage}{\columnwidth}
		\includegraphics[width=0.18\columnwidth]{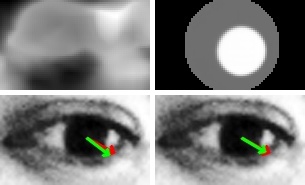} \hfill
		\includegraphics[width=0.18\columnwidth]{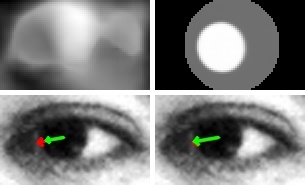} \hfill
		\includegraphics[width=0.18\columnwidth]{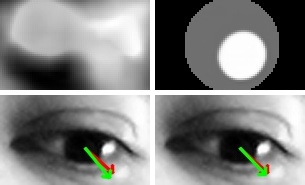} \hfill
		\includegraphics[width=0.18\columnwidth]{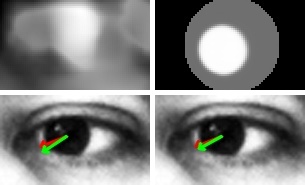} \hfill
		\includegraphics[width=0.18\columnwidth]{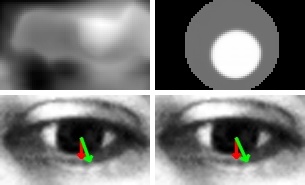}
				\subcaption{\scriptsize Intermediate representations and predictions from test samples without (left) and with (right) intermediate supervision}
	\end{minipage}
	\caption{
		Example of image representations learned by our architecture in the absence or presence of $\mathcal{L}_\mathrm{gazemap}$.
		Note that the pictorial representation is more consistent, and that the hourglass network is able to account for occlusions.
		Predicted gaze directions are shown in green, with ground-truth in red.
	}
	\label{fig:nogazemap}
\end{figure}

\section{Evaluations}
\label{sec:evaluations}
We perform our evaluations primarily on the MPIIGaze dataset, which consists of images taken of $15$ laptop users in everyday settings.
The dataset has been used as the standard benchmark dataset for unconstrained appearance-based gaze estimation in recent years \cite{Zhang2015CVPR,Wood2015ICCV,Wood2016ETRA,Shrivastava2017CVPR,Zhang2017CVPRW,Zhang2017PAMI}.
Our focus is on cross-person single-eye evaluations where $15$ models are trained per configuration or architecture in a leave-one-person-out fashion.
That is, a neural network is trained on $14$ peoples' data ($1500$ entries each from left and right eyes), then tested on the test set of the left-out person ($1000$ entries).
The mean over $15$ such evaluations is used as the final error metric representing cross-person performance.
As MPIIGaze is a dataset which well represents real-world settings, cross-person evaluations on the dataset is indicative of the real-world person-independence of a given model.

To further test the generalization capabilities of our method, we also perform evaluations on two additional datasets in this section: Columbia \cite{Smith2013UIST} and EYEDIAP \cite{FunesMora2014ETRA}, where we perform $5$-fold cross validation.
While Columbia displays large diversity between its $55$ participants, the images are of high quality, having been taken using a DSLR.
EYEDIAP on the other hand suffers from the low resolution of the VGA camera used, as well as large distance between camera and participant.
We select screen target (CS/DS) and static head pose sequences (S) from the EYEDIAP dataset, sampling every $15$ seconds from its VGA video streams (V). Training on moving head sequences (M) with just single eye input proved infeasible, with all models experiencing diverging test error during training.
Performance improvements on MPIIGaze, Columbia, and EYEDIAP would indicate that our model is robust to cross-person appearance variations and the challenges caused by low eye image resolution and quality.

In this section, we first evaluate the effect of our gazemap loss (Sec.~\ref{sec:eval_gazemap}), then compare the performance (Sec.~\ref{sec:eval_benchmark}) and robustness (Sec.~\ref{sec:eval_robustness}) of our approach against state-of-the-art architectures.

\subsection{Pictorial Representation (\emph{Gazemaps})}
\label{sec:eval_gazemap}

\newlength{\oldintextsep}
\setlength{\oldintextsep}{\intextsep}
\setlength{\intextsep}{0pt}
\begin{wraptable}{r}{4.7cm}
\vskip -0mm
\caption{Cross-person gaze estimation errors in the absence and presence of $\mathcal{L}_{gazemap}$, with DenseNet (k=32).}
\label{tab:eval_gazemap}
\centering
\renewcommand*{\arraystretch}{1.1}
\begin{tabular}{m{1.8cm} | >{\centering}m{1.1cm}>{\centering}m{1.1cm}}\hline
  \multirow{2}{*}{Dataset} &
  \multicolumn{2}{c}{$\mathcal{L}_\mathrm{gazemap}$}
    \tabularnewline
  & no & yes \tabularnewline
\hline
	MPIIGaze & 4.67  & \textbf{4.56}  \tabularnewline
	Columbia & 3.78  & \textbf{3.59}  \tabularnewline
	EYEDIAP  & 11.28 & \textbf{10.63} \tabularnewline
\hline
\end{tabular}
\end{wraptable}

We postulated in Sec.~\ref{sec:representation} that by providing a pictorial representation of 3D gaze direction that is visually similar to the input image, we could achieve improvements in appearance-based gaze estimation.
In our experiments we find that applying the gazemaps loss term generally offers performance improvements compared to the case where the loss term is not applied. This improvement is particularly emphasized when DenseNet growth rate is high (eg. $k=32$), as shown in Table~\ref{tab:eval_gazemap}.

By observing the output of the last hourglass module and comparing against the input images (Figure~\ref{fig:nogazemap}), we can confirm that even without intermediate supervision, our network learns to isolate the iris region, yielding a similar image representation of gaze direction across participants.
Note that this representation is learned only with the final gaze direction loss, $\mathcal{L}_\mathrm{gaze}$, and that blobs representing iris locations are not necessarily aligned with actual iris locations on the input images.
Without intermediate supervision, the learned minimal image representation may incorporate visual factors such as occlusion due to hair and eyeglases, as shown in Figure~\ref{fig:gazemap_comp}.

This supports our hypothesis that an intermediate representation consisting of an iris and eyeball contains the required information to regress gaze direction. However, due to the nature of learning, the network may also learn irrelevant details such as the edges of the glasses. Yet, by explicitly providing an intermediate representation in the form of gazemaps, we enforce a prior that helps the network learn the desired representation, without incorporating the previously mentioned unhelpful details. \setlength{\intextsep}{\oldintextsep}

\subsection{Cross-Person Gaze Estimation}
\label{sec:eval_benchmark}
We compare the cross-person performance of our model by conducting a leave-one-person-out evaluation on MPIIGaze and $5$-fold evaluations on Columbia and EYEDIAP.
In Section~\ref{sec:representation} we discussed that the mapping $k$ from gazemap to gaze direction should not require a complex architecture to model.
Thus, our DenseNet is configured with a low growth rate ($k=8$).
To allow fair comparison, we re-implement $2$ architectures for single-eye image inputs (of size $150\times 90$): AlexNet and VGG-16.
The AlexNet and VGG-16 architectures have been used in recent works in appearance-based gaze estimation and are thus suitable baselines \cite{Zhang2017CVPRW,Zhang2017PAMI}.
Implementation and training procedure details of these architectures are provided in supplementary materials.

\begin{table}[t]
\centering
\caption{
	Mean gaze estimation error in degrees for within-dataset cross-person $k$-fold evaluation.
	Evaluated on (a) MPIIGaze, (b) Columbia, and (c) EYEDIAP datasets.
\label{tab:eval_within}
}
\renewcommand{\arraystretch}{1.2}
\setlength{\tabcolsep}{5pt}
\begin{minipage}[c]{\columnwidth}
	\centering
	\label{tab:eval_mpi}
	\subcaption{MPIIGaze {\scriptsize($15$-fold)}}
		\begin{tabular}{l|c|c|c|c|c|c|c}
		\hline
		Model                        &
		kNN \cite{Zhang2017PAMI}     &
		RF \cite{Zhang2017PAMI}      &
		\cite{Zhang2015CVPR}         &
		AlexNet                      &
		VGG-16                       &
		GazeNet \cite{Zhang2017PAMI} &
		\textbf{ours}                \\
		\hline
		\# params & $0$   & -     & $1.8$M & $86$M & $158$M & $90$M & $0.7$M \\
		Inputs    & e + h & e + h & e + h  & e     & e      & e + h & e      \\
		\hline
		Error     & $7.2$ & $6.7$ & $6.3$  & $5.7$ & $5.4$  & $5.5$ & $\mathbf{4.5}$ \\
		\hline
	\end{tabular}
\end{minipage}
\hfill
\begin{minipage}{0.47\columnwidth}
	\centering
	\label{tab:eval_columbia}
	\subcaption{Columbia {\scriptsize($5$-fold)}}
		\begin{tabular}{l|c|c|c}
		\hline
		Model & AlexNet & VGG-16 & \textbf{ours}  \\
		\hline
		Error & $4.2$   & $3.9$  & $\mathbf{3.8}$ \\
		\hline
	\end{tabular}
\end{minipage}
\hfill
\begin{minipage}{0.47\columnwidth}
	\centering
	\label{tab:eval_eyediap}
	\subcaption{EYEDIAP {\scriptsize($5$-fold)}}
		\begin{tabular}{l|c|c|c}
		\hline
		Model & AlexNet & VGG-16 & \textbf{ours} \\
		\hline
		Error & $11.5$  & $11.2$ & $\mathbf{10.3}$ \\
		\hline
	\end{tabular}
\end{minipage}
\vskip 2mm
{\scriptsize where e: single-eye, h: head pose (pitch, yaw)}
\end{table}

\begin{figure}[t]
	\begin{minipage}{\columnwidth}
		\centering
		\includegraphics[width=0.32\columnwidth]{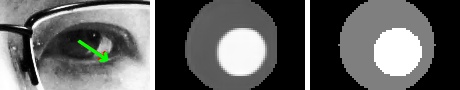}\hfill
		\includegraphics[width=0.32\columnwidth]{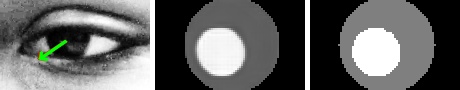}\hfill
		\includegraphics[width=0.32\columnwidth]{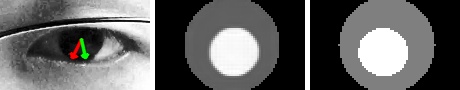}\\
		\includegraphics[width=0.32\columnwidth]{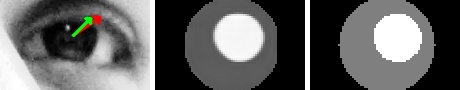}\hfill
		\includegraphics[width=0.32\columnwidth]{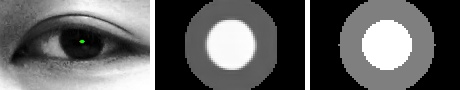}\hfill
		\includegraphics[width=0.32\columnwidth]{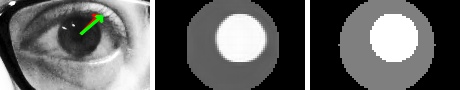}\\
		\includegraphics[width=0.32\columnwidth]{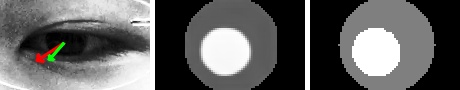}\hfill
		\includegraphics[width=0.32\columnwidth]{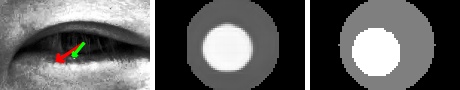}\hfill
		\includegraphics[width=0.32\columnwidth]{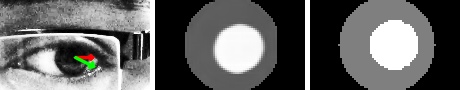}
		\vskip -1mm
		\subcaption{Columbia}
	\end{minipage}
	\vskip 1mm
	\begin{minipage}{\columnwidth}
		\centering
		\includegraphics[width=0.32\columnwidth]{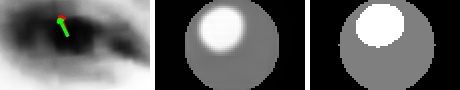}\hfill
		\includegraphics[width=0.32\columnwidth]{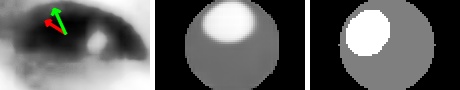}\hfill
		\includegraphics[width=0.32\columnwidth]{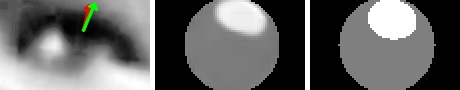}\\
		\includegraphics[width=0.32\columnwidth]{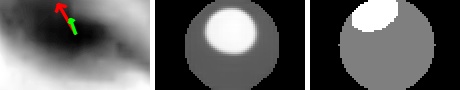}\hfill
		\includegraphics[width=0.32\columnwidth]{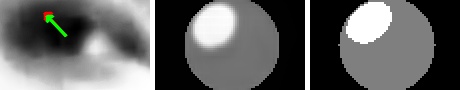}\hfill
		\includegraphics[width=0.32\columnwidth]{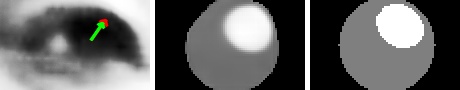}\\
		\includegraphics[width=0.32\columnwidth]{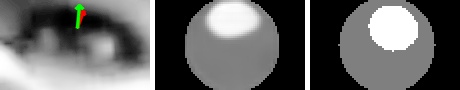}\hfill
		\includegraphics[width=0.32\columnwidth]{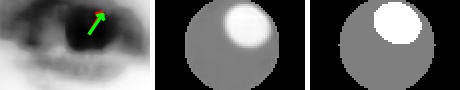}\hfill
		\includegraphics[width=0.32\columnwidth]{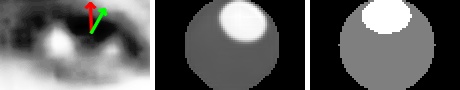}
		\vskip -1mm
		\subcaption{EYEDIAP}
	\end{minipage}
	\caption{
		Gazemap predictions (middle) on Columbia and EYEDIAP datasets with ground-truth (red) and predicted (green) gaze directions visualized on input eye images (left).
		Ground-truth gazemaps are shown on the far-right of each triplet.
	}
	\label{fig:columbia_eyediap}
\end{figure}

In MPIIGaze evaluations (Table ~\ref{tab:eval_mpi}a), our proposed approach outperforms the current state-of-the-art approach by a large margin, yielding an improvement of $1.0^\circ$ ($5.5^\circ\rightarrow 4.5^\circ=18.2\%$).
This significant improvement is in spite of the reduced number of trainable parameters used in our architecture (90M vs 0.7M).
Our performance compares favorably to that reported in \cite{Zhang2017CVPRW} ($4.8^\circ$) where full-face input is used in contrast to our single-eye input.
While our results cannot directly be compared with those of \cite{Zhang2017CVPRW} due to the different definition of gaze direction (face-centred as opposed to eye centred), the similar performance suggests that eye images may be sufficient as input to the task of gaze direction estimation.
Our approach attains comparable performance to models taking face input, and uses considerably less parameters than recently introduced architectures ($129$x less than GazeNet).

We additionally evaluate our model on the Columbia Gaze and EYEDIAP datasets in Table~\ref{tab:eval_columbia}b and Table~\ref{tab:eval_eyediap}c respectively.
While high image quality results in all three methods performing comparably for Columbia Gaze, our approach still prevails with an improvement of $0.4^\circ$ over AlexNet.
On EYEDIAP, the mean error is very high due to the low resolution and low quality input.
Note that there is no head pose estimation performed, with only single eye input being relied on for gaze estimation.
Our gazemap-based architecture shows its strengths in this case, performing $0.9^\circ$ better than VGG-16 - a $8\%$ improvement.
Sample gazemap and gaze direction predictions are shown in Figure~\ref{fig:columbia_eyediap} where it is evident that despite the lack of visual detail, it is possible to fit gazemaps to yield improved gaze estimation error.

By evaluating our architecture on $3$ different datasets with different properties in the cross-person setting, we can conclude that our approach provides significantly higher generalization capabilities compared to previous approaches.
Thus, we bring gaze estimation closer to direct real-world applications.

\subsection{Robustness Analysis}
\label{sec:eval_robustness}
\begin{figure}
\begin{minipage}{0.48\columnwidth}
	\centering
	\includegraphics[width=\columnwidth]{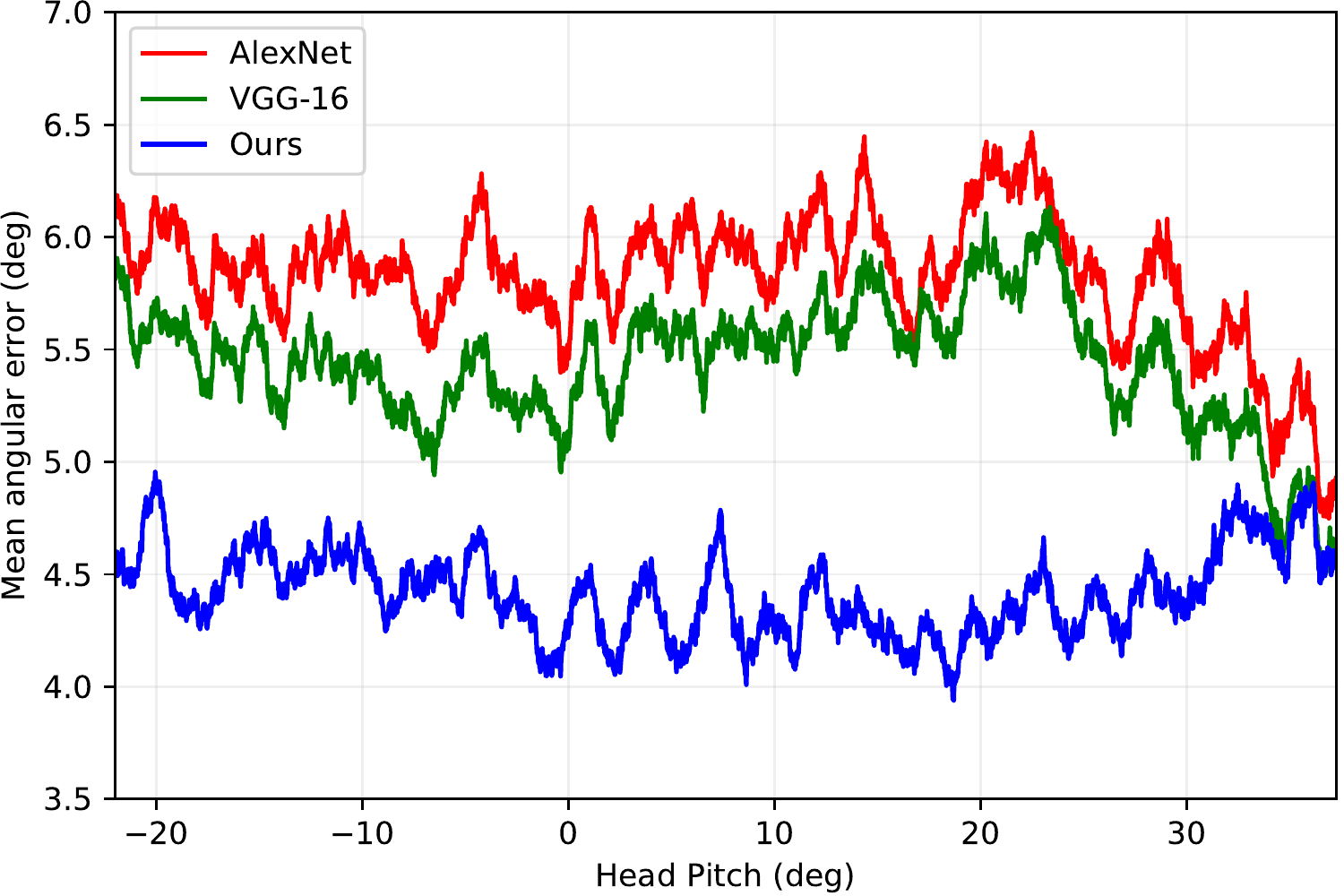}
	\subcaption{}\end{minipage}
\hfill
\begin{minipage}{0.48\columnwidth}
	\centering
	\includegraphics[width=\columnwidth]{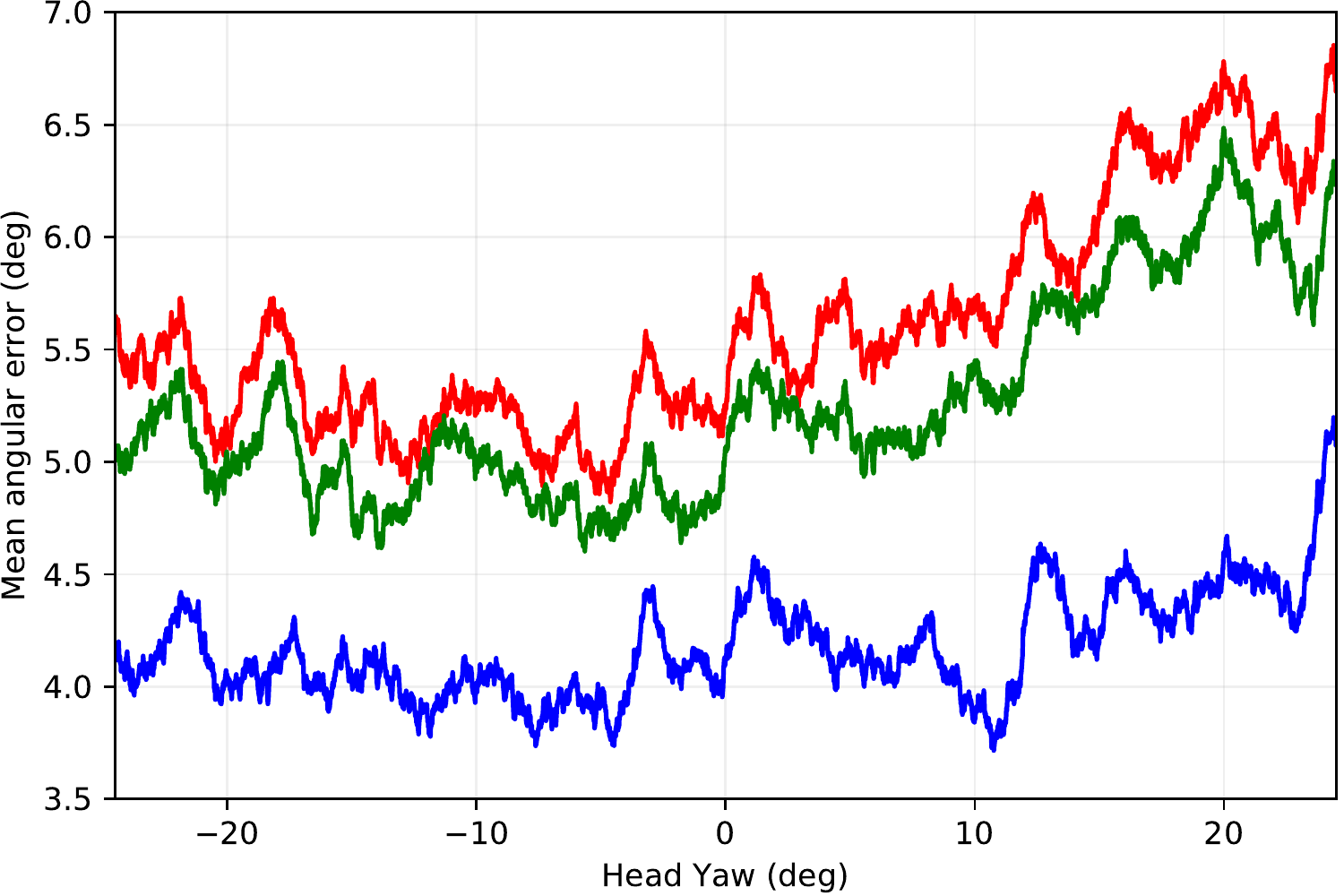}
	\subcaption{}\end{minipage}
\label{fig:head_pitch_yaw}
\begin{minipage}{0.48\columnwidth}
	\centering
	\includegraphics[width=\columnwidth]{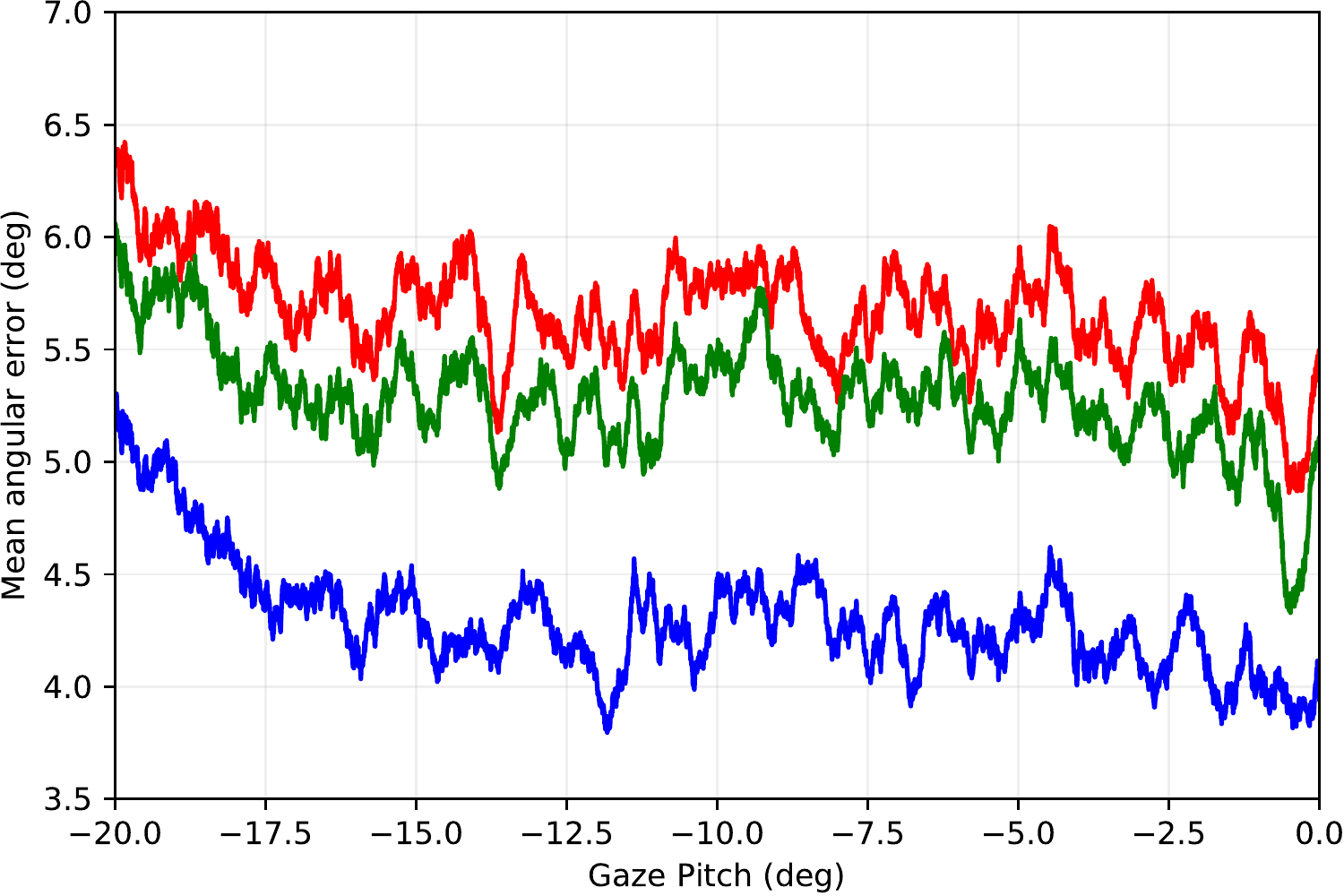}
	\subcaption{}\end{minipage}
\hfill
\begin{minipage}{0.48\columnwidth}
	\centering
	\includegraphics[width=\columnwidth]{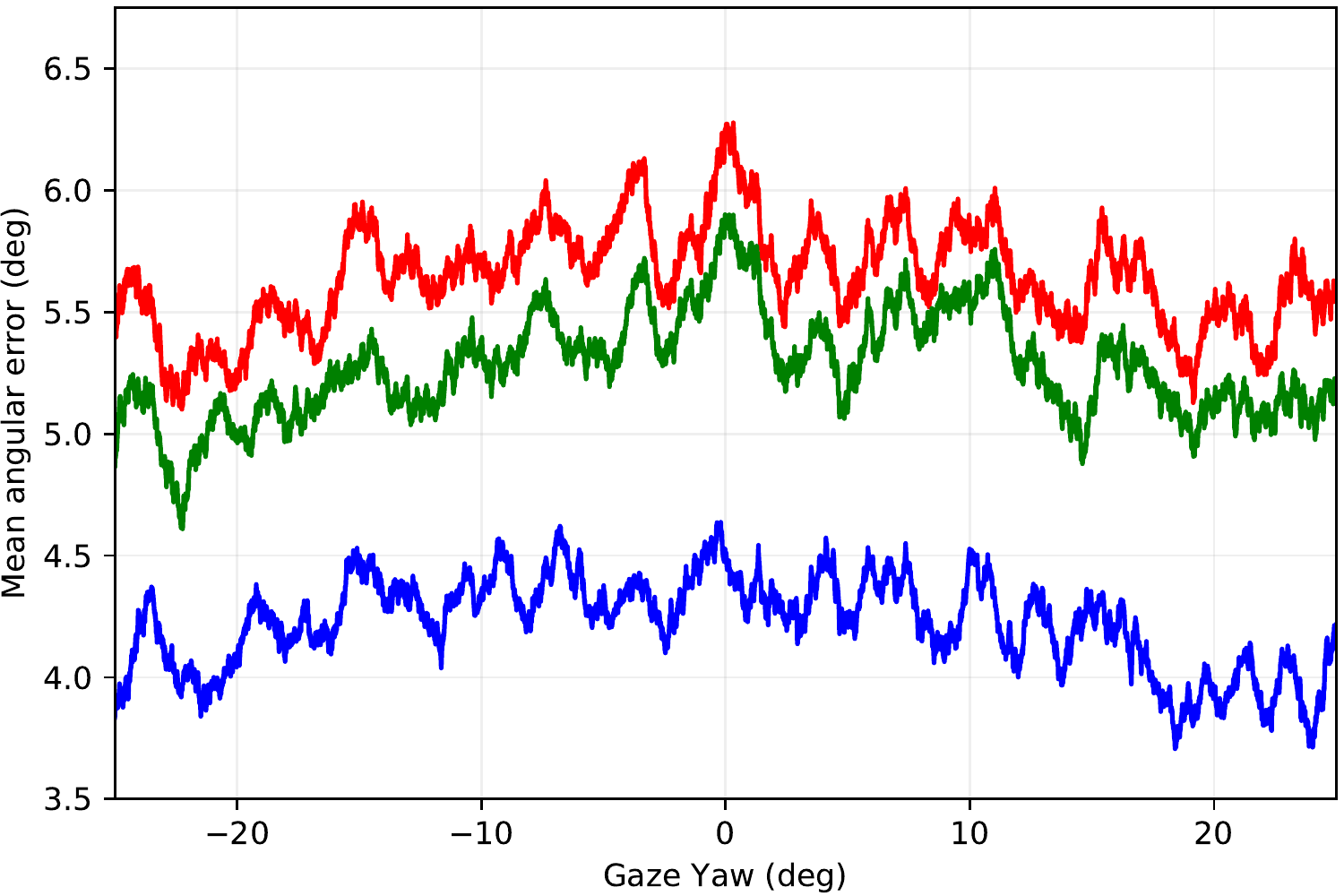}
	\subcaption{}\end{minipage}
\begin{minipage}{0.48\columnwidth}
	\centering
	\includegraphics[width=\columnwidth]{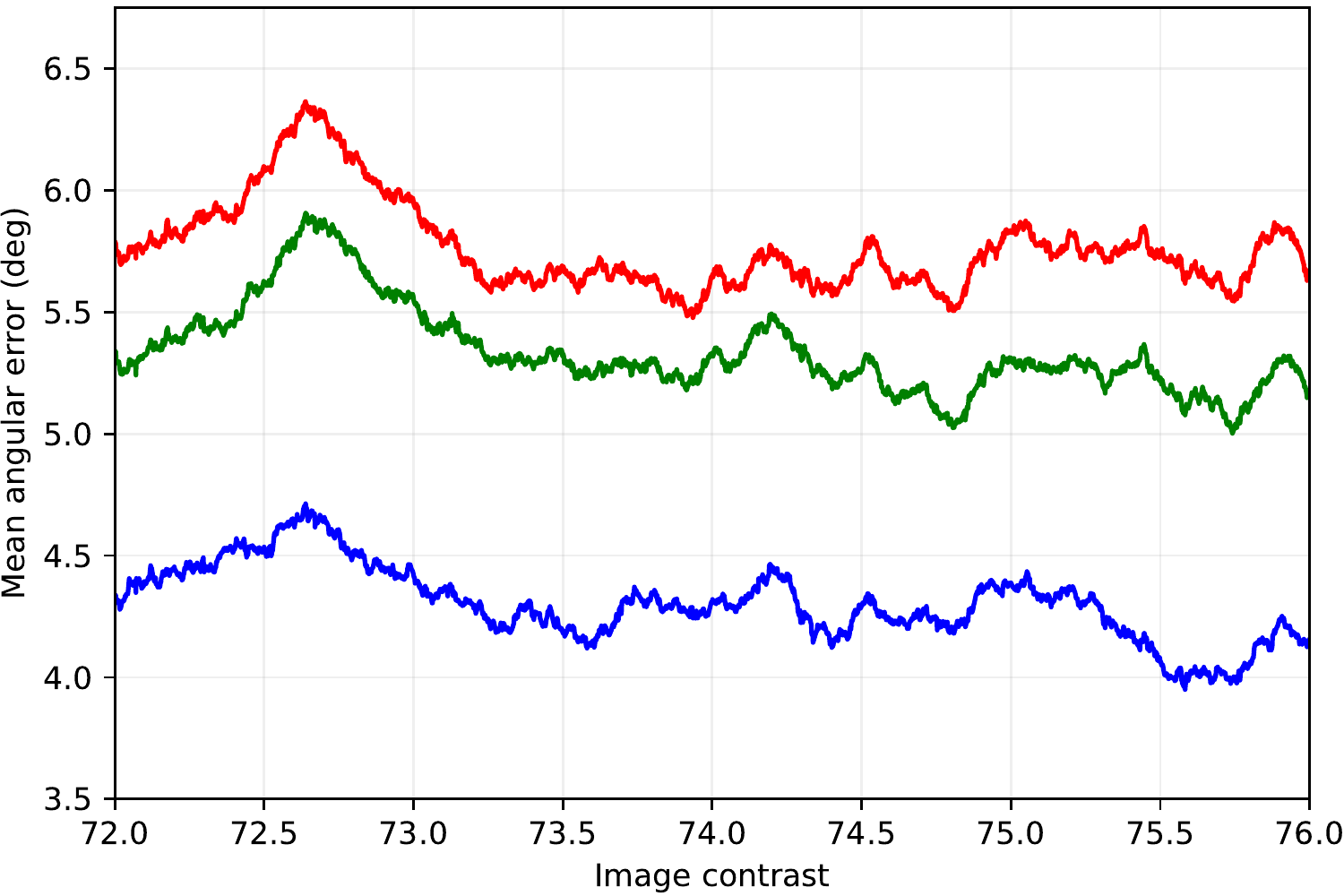}
	\subcaption{}\end{minipage}
\hfill
\begin{minipage}{0.48\columnwidth}
	\centering
	\includegraphics[width=\columnwidth]{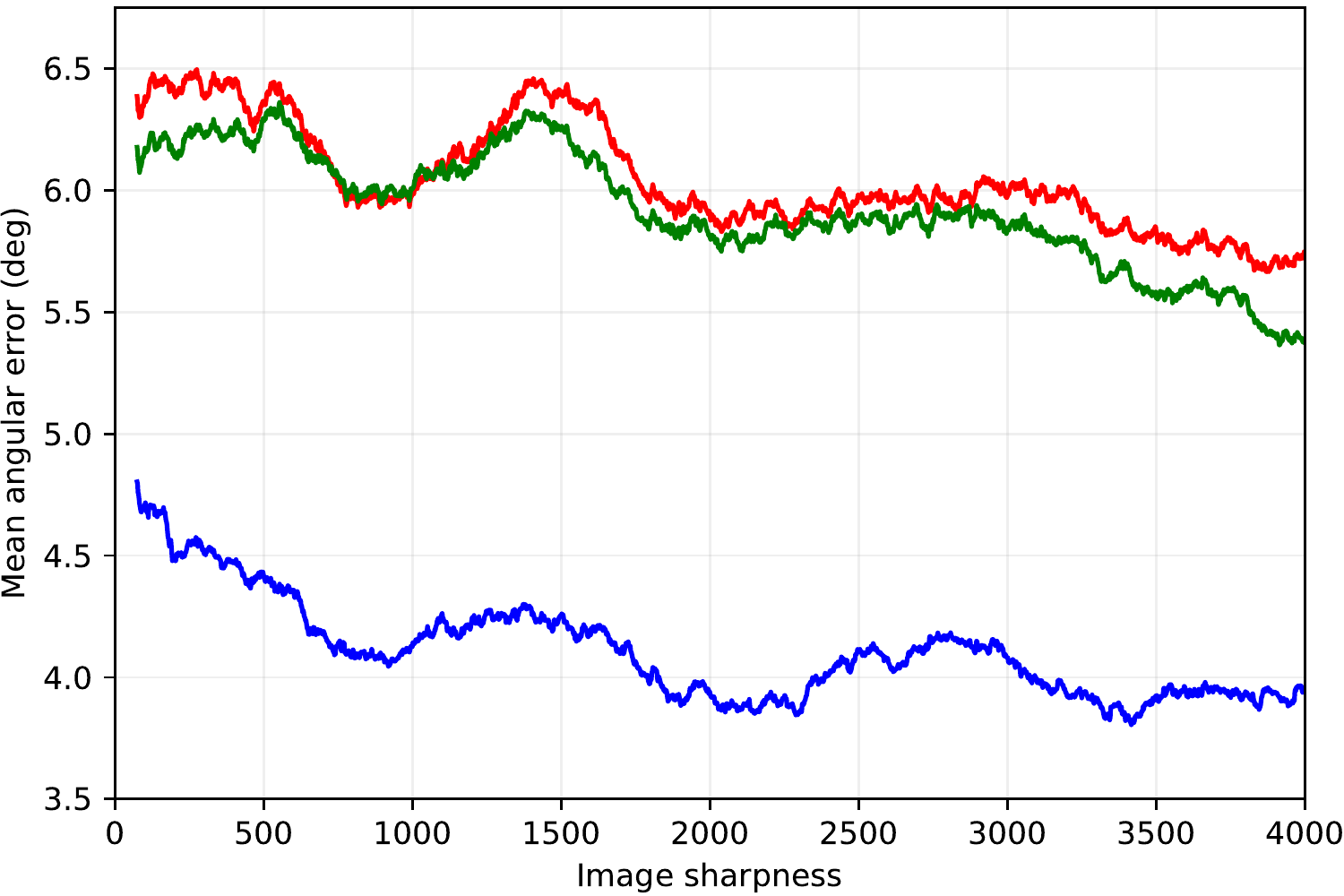}
	\subcaption{}\end{minipage}
\caption{Robustness of AlexNet (red), VGG-16 (green), and our approach (blue) to different head pose (top), gaze direction (middle), and image quality (bottom). The lines are a moving average.}
\label{fig:robustness_analysis}
\end{figure}

In order to shed more light onto our models' performance, we perform an additional robustness analysis. More concretely, we aim to analyze how our approach performs under difficult and challenging situations, such as extreme head pose and gaze direction. In order to do so, we evaluate a moving average on the output of our within-MPIIGaze evaluations, where the $y$-values correspond to the mean angular error and the $x$-values take \emph{one} of the following factor of variations: head pose (pitch \& yaw), gaze direction (pitch \& yaw). Additionally, we also consider image quality (contrast \& sharpness) as a qualitative factor.
In order to isolate each factor of variation from the rest, we evaluate the moving average only on the points whose remaining factors are close to its median value. Intuitively, this corresponds to data points where the person moves only in one specific direction, while staying at rest in all of the remaining directions.
This is not the case for image quality analysis, where all data points are used.
Figure~\ref{fig:robustness_analysis} plots the mean angular error as a function of different movement variations and image qualities. The top row corresponds to variation along the head pose, the middle along gaze direction and the bottom to varying image quality. In order to calculate the image contrast, we used the RMS contrast metric whereas to compute the sharpness, we employ a Laplacian-based formula as outlined in \cite{Pech2000ICPR}. Both metrics are explained in supplementary materials.
The figure shows that we \emph{consistently} outperform competing architectures for extreme head and gaze angles.
Notably, we show more consistent performance in particular over large ranges of head pitch and gaze yaw angles.
In addition, we surpass prior works on images of varying quality, as shown in Figures \ref{fig:robustness_analysis}e and \ref{fig:robustness_analysis}f.

\section{Conclusion}
Our work is a first attempt at proposing an explicit prior designed for the task of gaze estimation with a neural network architecture.
We do so by introducing a novel pictorial representation which we call gazemaps.
An accompanying architecture and training scheme using intermediate supervision naturally arises as a consequence, with a fully convolutional architecture being employed for the first time for appearance-based eye gaze estimation.
Our gazemaps are anatomically inspired, and are experimentally shown to outperform approaches which consist of significantly more model parameters and at times, more input modalities.
We report improvements of up to $18\%$ on MPIIGaze along with improvements on additional two different datasets against competitive baselines.
In addition, we demonstrate that our final model is more robust to various factors such as extreme head poses and gaze directions, as well as poor image quality compared to prior work.

Future work can look into alternative pictorial representations for gaze estimation, and an alternative architecture for gazemap prediction. Additionally, there is potential in using synthesized gaze directions (and corresponding gazemaps) for unsupervised training of the gaze regression function, to further improve performance.
 \section*{Acknowledgements}
This work was supported in part by ERC Grant OPTINT (StG-2016-717054).
We thank the NVIDIA Corporation for the donation of GPUs used in this work.

\bibliographystyle{splncs04}
\bibliography{bibliography}

\renewcommand\thesection{\Alph{section}}
\setcounter{section}{0}
\section{Baseline Architectures}

The state-of-the-art CNN architecture for appearance-based gaze estimation is based on a lightly modified VGG-16 architecture \cite{Zhang2017PAMI}, with mean cross-person gaze estimation error of $5.5^\circ$ on the MPIIGaze dataset \cite{Zhang2015CVPR}.
We compare against a standard VGG-16 architecture \cite{Simonyan15ICLR} and an AlexNet architecture \cite{Krizhevsky12NIPS} which has been the standard architecture for gaze estimation in many works \cite{Krafka2016CVPR,Zhang2017CVPRW}.
The specific architectures used as baseline are described in Table~\ref{tab:baseline_archs}.

Both models are trained with a batch size of $32$, learning rate of $5\times10^{-5}$ and $L_2$ weights regularization coefficient of $10^{-4}$, using the Adam optimizer \cite{Kingma2014arXiv}.
Learning rate is multiplied by $0.1$ every $5,000$ training steps, and slight data augmentation is performed in image translation and scale.
 \section{Image metrics}
In this section we describe the image metrics used for the robustness plots concerning image quality (Figures 6e and 6f in paper).
\subsection{Image contrast}
The root mean contrast is defined as the standard deviation of the pixel intensities:
$$
\text{RMC} = \sqrt{\frac{1}{MN}\sum_{i=1}^{N-1}\sum_{j=1}^{M-1}(I_{ij} - \bar{I})^2}
$$
where $I_{ij}$ is the value of the image $I\in M \times N$ at location $(i,j)$ and $\bar{I}$ is the average intensity of all pixel values in the image.
\subsection{Image sharpness}
In order to have a sharpness-based metric, we calculate the variance of the image $I$ after having convolved it with a Laplacian, similar to \cite{Pech2000ICPR}. This corresponds to an approximation of the second derivative, which is computed with the help of the following mask:
$$
L = \frac{1}{6} \begin{bmatrix}
    0 & -1 & 0 \\
    -1 & 4 & -1 \\
    0 & -1 & 0
\end{bmatrix}
$$
After convolving $I$ with $L$, we compute the standard deviation of the resulting image $I_L$ to get the image sharpness (IS) metric:
$$
\text{IS} = \sigma(I_L)
$$
 \begin{table}
\caption{
	Configuration of CNNs used as baseline for gaze estimation.
	The style of \cite{Simonyan15ICLR} is followed where possible.
	$s$ represents stride length, $p$ dropout probability, and conv9-96 represents a convolutional layer with kernel size $9$ and $96$ output feature maps.
	maxpool3 represents a max-pooling layer with kernel size $3$.
}
\label{tab:baseline_archs}
\renewcommand{\arraystretch}{1.2}
\setlength{\tabcolsep}{5pt}
\begin{minipage}[t]{0.5\columnwidth}
	\subcaption{AlexNet}
	\centering
	\begin{tabular}{c}
		\hline
		input ($150\times90$ eye image) \\
		\hline
		conv9-96 ($s=2$) \\
		local response norm. \\
		\hline
		maxpool3 ($s=2$) \\
		\hline
		conv5-256 ($s=1$) \\
		local response norm. \\
		\hline
		maxpool3 ($s=2$) \\
		\hline
		conv3-384 ($s=1$) \\
		conv3-384 ($s=1$) \\
		conv3-256 ($s=1$) \\
		\hline
		maxpool3 ($s=2$) \\
		\hline
		FC-4096 \\
		dropout ($p=0.5$) \\
		FC-4096 \\
		dropout ($p=0.5$) \\
		FC-2 \\
		\hline
	\end{tabular}
\end{minipage}
\begin{minipage}[t]{0.5\columnwidth}
	\subcaption{VGG-16}
	\centering
	\begin{tabular}{c}
		\hline
		input ($150\times90$ eye image) \\
		\hline
		conv3-64 \\
		conv3-64 \\
		\hline
		maxpool2 ($s=1$) \\
		\hline
		conv3-128 \\
		conv3-128 \\
		\hline
		maxpool2 ($s=2$) \\
		\hline
		conv3-256 \\
		conv3-256 \\
		conv3-256 \\
		\hline
		maxpool2 ($s=2$) \\
		\hline
		conv3-512 \\
		conv3-512 \\
		conv3-512 \\
		\hline
		maxpool2 ($s=2$) \\
		\hline
		conv3-512 \\
		conv3-512 \\
		conv3-512 \\
		\hline
		maxpool2 ($s=2$) \\
		\hline
		dropout ($p=0.5$) \\
		FC-4096 \\
		dropout ($p=0.5$) \\
		FC-4096 \\
		FC-2 \\
		\hline
	\end{tabular}
\end{minipage}
\end{table}

\end{document}